\documentclass{article}

\usepackage{PRIMEarxiv}
\usepackage{float}
\usepackage{amsmath}

\usepackage{subcaption}
\usepackage{enumitem}
\usepackage[utf8]{inputenc} 
\usepackage[T1]{fontenc}    
\usepackage{hyperref}       
\usepackage{url}            
\usepackage{booktabs}       
\usepackage{amsfonts}       
\usepackage{nicefrac}       
\usepackage{microtype}      
\usepackage{lipsum}
\usepackage{fancyhdr}       
\usepackage{graphicx}       
\graphicspath{{media/}}     

\title{Optimal Kinematic Synthesis and
Prototype Development of Knee
Exoskeleton
}

\author{
  Shashank Mani Gautam \\
  IIT Ropar \\
  Punjab, India\\
  \texttt{shashankmani2020@gmail.com} \\
   \And
  Dr. Ekta Singla \\
  IIT Ropar \\
  Punjab, India \\
  \texttt{ekta@iitrpr.ac.in} \\
  \AND
  Dr. Ashish Singla \\
  TIET Patiala \\
  Punjab, India \\
  \texttt{ashish.singla@thapar.edu} \\
}

\begin{document}
\maketitle

\begin{abstract}
The range of rotation (RoR) in a knee exoskeleton is a critical factor in rehabilitation, as it directly influences joint mobility, muscle activation, and recovery outcomes. A well-designed RoR ensures that patients achieve near-natural knee kinematics, which is essential for restoring gait patterns and preventing compensatory movements. This paper presents optimal design of one degree of freedom knee exoskeleton.
In kinematic analysis, the existing design being represented by nonlinear and nonconvex mathematical functions. To obtain feasible and optimum measurement of the links of knee exoskeleton, an optimization problem is formulated based on the kinematic analysis and average human's leg measurement. The optimized solution increases the range of motion of knee exoskeleton during sit to stand motion by $24 \%$ as compared with inspired design. 
Furthermore, misalignment study is conducted by comparing the trajectory of human's knee and exoskeleton's knee during sit to stand motion. The joint movement is calculated using marker and camera system. 
Finally, a prototype of the knee joint exoskeleton is being developed based on optimal dimensions which validate the maximum range of motion achieved during simulation.
\end{abstract}


\section{Introduction}
Patients with mobility impairments resulting from stroke, osteoarthritis, or spinal cord injuries require rehabilitation for post-recovery. Knee exoskeleton have emerged as transformative tools in rehabilitation by providing mechanical support and guided movement, these devices help restore natural gait patterns, reduce muscle atrophy, and accelerate recovery \cite{veneman2007design}.  However, their effectiveness depends on the refined mechanical design particularly in replicating the biomechanics of the human knee.

Among various designs, single degree-of-freedom (DoF) exoskeletons offer significant advantages over their multi-DoF counterparts, especially in terms of simplicity, reliability, and wearability. Unlike complex multi-joint exoskeletons that require advanced control systems, higher energy consumption and are susceptible to misalignment, single-DoF designs focus solely on flexion and extension, which are the primary motions needed for walking, stair climbing, and transitioning from sitting to standing. This simplified approach reduces mechanical complexity, minimizes weight, and enhances user comfort, making single-DoF exoskeletons more practical for daily rehabilitation use \cite{xiao2020bionic}.

However, a critical challenge in single-DoF knee exoskeletons is ensuring that their joint mechanics accurately replicate the natural kinematics of the human knee. This is where a four-bar linkage mechanisms is more useful. Unlike traditional two-bar hinge structures that have a fixed Instantaneous Center of Rotation (ICR), the four-bar mechanism allows the ICR to move in a J-shaped trajectory, similar to the human knee joint \cite{dollar2008lower}. By incorporating a movable pivot system—comprising femoral, tibial, and two connecting links—the four-bar design dynamically adjusts its rotation center to closely match the knee’s natural ICR trajectory.

To ensure this, the most common way is to find the optimal solutions of given design. In the field of exoskeleton, most design problem are complex. Depending on the objective and design of the exoskeleton, different techniques have been developed to get an approximate solution. Researchers have used different optimization methods to mechanically design exoskeleton which fulfill their specific goal.
Zakaryan et al.\cite{zakaryan2021bio} applied a weighted-sum Differential Evolution (DE) approach \cite{rodriguez2016multiobjective} to optimize joint weights and link dimensions for an arm exoskeleton, aiming to emulate human muscle behavior by minimizing mass, peak cable tensions, and imbalances between agonist and antagonist tensions. 
Similarly, Tian et al.\cite{tian2017mechanism} used Particle Swarm Optimization (PSO) to optimize link lengths for enhanced force transmission and sit-to-stand support in a leg exoskeleton. Du et al.\cite{du2020mechanical} also employed PSO to fine-tune rope routing and link lengths in a cable-driven arm mechanism, focusing on maximizing torque and reducing misalignment.

Deboer et al. \cite{deboer2022discrete} used NSGA-II to find optimal link lengths, spring stiffness, angles, and displacements for a leg exoskeleton, reducing peak power and total length. Paez et al. \cite{paez2022mobility} optimized link lengths and joint locations for a knee exoskeleton to minimize joint load and torque deviation, promoting natural motion.
Rituraj et al. \cite{rituraj2022hydraulic} determined the optimal link lengths and angles for their assistive/rehabilitative knee exoskeleton for rehabilitation and assistive use. The optimal configuration minimizes the maximum distance between the actuators and the load on the device. 
McDaid \cite{mcdaid2017design} used WBGA to find optimal link lengths for a leg exoskeleton, aiming to maximize workspace, avoid singularities, and minimize size. Asker et al. \cite{asker2021multi} employed WBGA to optimize link lengths for knee exoskeletons, focusing on maximizing force transmission and minimizing misalignment.

Researcher are still working on the optimization of mechanical design of exoskeleton and as per author’s knowledge and given literature review, no work has been found where a knee joint exoskeleton has been optimized using interior point technique. In this paper, authors have developed an optimization algorithm based on Interior Point Method optimization technique and performed feasibility check using graphical simulation to mechanical model the Linear Actuator based Knee joint exoskeleton for maximum range of motion during Sit to Stand (STS) motion. Figure \ref{exo_cad} shows the knee exoskeleton inspired by Jain et al. \cite{jain2022linear} and Kim et al. \cite{kim2014design}. The proposed design is based on a planar four-bar mechanism actuated by a linear DC motor. It consist of universal design that allows it to be worn by anyone, irrespective of their leg length. The exoskeleton's fifth joint is positioned at the wearer's knee joint, while the shank length can be adjusted to fit the wearer's calf length. 
The paper is structured as follows: Section 2 describe the architecture of existing exoskeleton and  its kinematic modelling. Section 3 and 4 discusses the  formulation of objective functions and optimization results for the STS motion. Section 5 and 6 contains the graphical simulation of optimized exoskeleton and gait analysis. Finally, Section 7 and 8 presents the validation of the optimal result by prototype development and conclusion.

\section{ Kinematic Modeling}


\begin{figure}[h]
    \centering
    \begin{subfigure}{0.5 \textwidth}
        \centering
        \includegraphics[width = \textwidth]{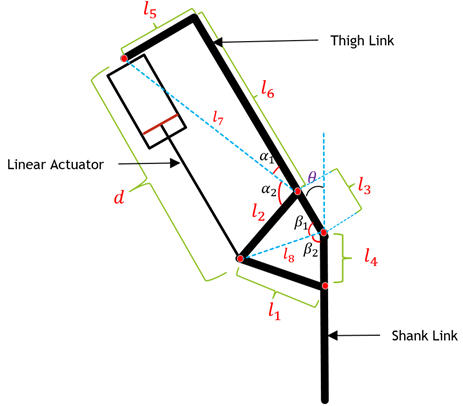}
        \caption{Schematic diagram of existing knee joint exoskeleton}
        \label{Geometric Diagram}
    \end{subfigure}
    \hfill
    \begin{subfigure}{0.4\textwidth}
        \centering
        \includegraphics[width = 0.8 \textwidth]{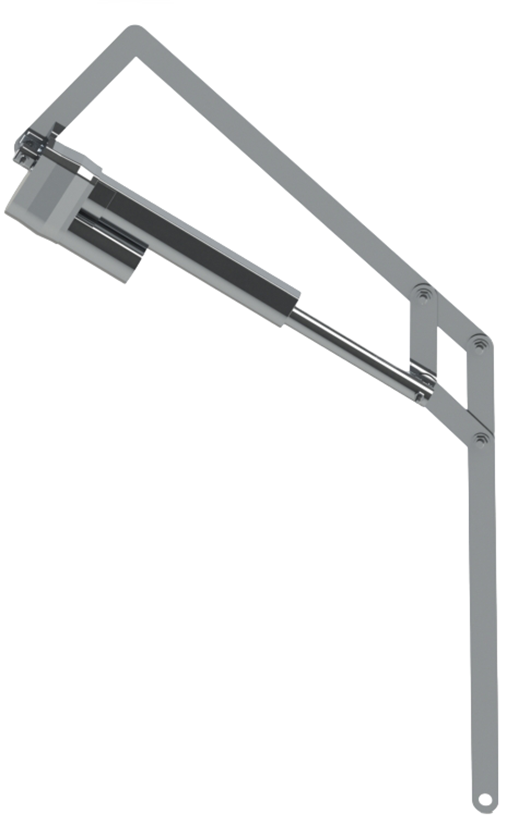}
        \caption{CAD model of knee exoskeleton.}
        \label{exo_cad}
    \end{subfigure}
    \caption{Linear Actuator based Knee Exoskeleton (LAKE) }
    \label{fig:comparison}
\end{figure}
Figure \ref{Geometric Diagram} describes a robotic exoskeleton intended to support human mobility. The core component of the exoskeleton is a four-bar mechanism. This mechanism is composed of four rigid links, namely $l_1$, $l_2$, $l_3$, and $l_4$ which are linked by four pin joints, forming a quadrilateral shape. The exoskeleton's motion is driven by a linear DC motor, which, when activated, propels a piston that sets in motion a slewing bar ($l_2$), causing the thigh link ($l_6$) to pivot around a specific point (5) while the shank link ($l_6$) remains attached to the ground (ankle).
The design of the exoskeleton is based on a crank-rocker mechanism that meets Grashof's criteria. This mechanism is characterized by a sum of the shortest and longest links ($l_3$ and $l_4$) that is less than the sum of the other two links ($l_1$ and $l_2$). Additionally, the link adjacent to the shortest link ($l_4$) remains fixed during STS motion. As a result of these features, the shortest link ($l_3$) acts as a crank, driving the mechanism's motion. Meanwhile, the link next to the fixed link ($l_1$) behaves as a rocker, oscillating back and forth. The coordinated interplay between the links and the motor enables the exoskeleton to move its limbs, thus emulating the natural gait of humans.
The mathematical formulation based on the exoskeleton in Fig \ref{exo_cad} is detailed here keeping $ \theta $ as the knee joint angle - representing the degree of freedom of the system.
\begin{align}
&\theta(t) = 180^{\circ}-\beta_1(t)-\beta_2(t)\\
&\beta_1(t)=\cos ^{-1}\left(\left(l_8^2+l_3^2-l_2^2\right) / 2 l_8 l_3\right) \\
&\beta_2(t)=\cos ^{-1}\left(\left(l_8^2+l_4^2-l_1^2\right) / 2 l_8 l_4\right) \\
& l_8=\left(l_2^2+l_3^2-2 l_2 l_3 \cos \left(180-\alpha_1-\alpha_2\right)\right)^{1 / 2} \\
& \alpha_1=\tan ^{-1}\left(l_5 / l_6\right) \\
& \alpha_2=\cos ^{-1}\left(\left(l_7^2+l_2^2-l_9^2\right) / 2 l_7 l_2\right) 
\end{align}

where $d$ is the stroke length of linear actuator. 
$\alpha_1$ is the angle between link $l_5$ and $l_6$.
$\alpha_2$  is the angle between link $l_7$ and $l_2$.
$\beta_1$ is the angle between link $l_3$ and $l_8$.
$\beta_2$ is the angle between link $l_8$ and $l_4$ and both $\beta$ angle changes according to design link variables explained in Section 3.2.

\section{ Design Optimization}
\subsection{Problem Formulation}
The angle of the knee joint in an exoskeleton is governed by a linear actuator. The maximum angle of rotation achievable by the knee joint is contingent on the minimum stroke length possible with the linear actuator. Based on Jain et al \cite{jain2022linear}, this minimum stroke length is taken as 242 mm. The optimization of the six remaining links that indirectly impact the knee joint's range of motion is essential to increase the latter. The main objective is to maximize the function $ F(x) $ defined in equation (\ref{eq7}) with design variable represented in equation  (\ref{eq8}).

\begin{align}
    \ F(x) : \theta = 180^{\circ}-\beta_1(t)-\beta_2(t)
    \label{eq7}
\end{align}

\subsection{Design Variables}
The exoskeleton consists of 6 independent links whose length dimensions are considered as design variables in this problem. This selection provides a larger domain for acquiring optimal solution. Also the width of all the links are assumed to be same.

\begin{align}
    \textbf{X} = \{l_1,l_2,l_3,l_4,l_5,l_6 \}^t
    \label{eq8}
\end{align}
\\
Based on these six variables, the objection function can be represented as

\begin{align}
max \; f = \theta(\textbf{X},d) \notag
\end{align}
\\
where
$d$ = $d_{min} = 242 \; mm $ \cite{jain2022linear}

\subsection{Constraints}
The proposed exoskeleton comprises of 6 links, and to obtain the optimal measurement for maximum range of motion for every user, constraints play a crucial role in the optimization. For optimal solution of objective function in Section 3.2, the constraints are proposed based on mechanical and human knee restrictions in real world scenario.
\begin{enumerate}
    \item The four bar linkage consist of $l_1$, $l_2$, $l_3$,$l_4$ link. The four bar mechanism should follow the Grashof’s law i.e. the sum of shortest and largest link should be less than the sum of other two links. i.e. \quad $l_3+l_4-l_2-l_1 < 0$    
    \item The above constrains follow the Grashof’s law but while designing there is no information which is the shortest link in all four links. Therefore, $l_3-l_2 < 0$ and $l_3-l_1 < 0$. 
    \item Similarly, there is no information of longest link also. Therefore, there are addition of given constrains \quad $l_3-l_4 < 0$, $l_2-l_4 < 0$, $l_1-l_4 < 0$
    \item Since the width of the human thigh is always less than its length, a constraint is imposed as $l_1-l_4 < 0$, $l_5-l_6 < 0$, $l_4-l_6 < 0$ accurately reflect this anatomical relationship in the optimization problem.
    \item Since, the optimization is purely based on geometrical approach, it is important to give constrains to knee joint angle up to the feasible limit.To achieve this, a singularity constraint has been added into the problem. Figure \ref{singularity} shows the position of links in the triangle when reaching at the maximum knee joint angle of exoskeleton. The further increment of angle reduces the angle between $l_7$ and $d$. When the angle between link $d$ and $l_2$ becomes  $180^\circ$, then it loses its one degree of freedom which convert this mechanism into structure. Therefore, the mechanism become structure if it satisfies the constrains $l_7$ = $l_2$ + $d_{min}$. 
 
    \begin{figure}[h]
        \centering
        \includegraphics[width = 0.4 \textwidth]{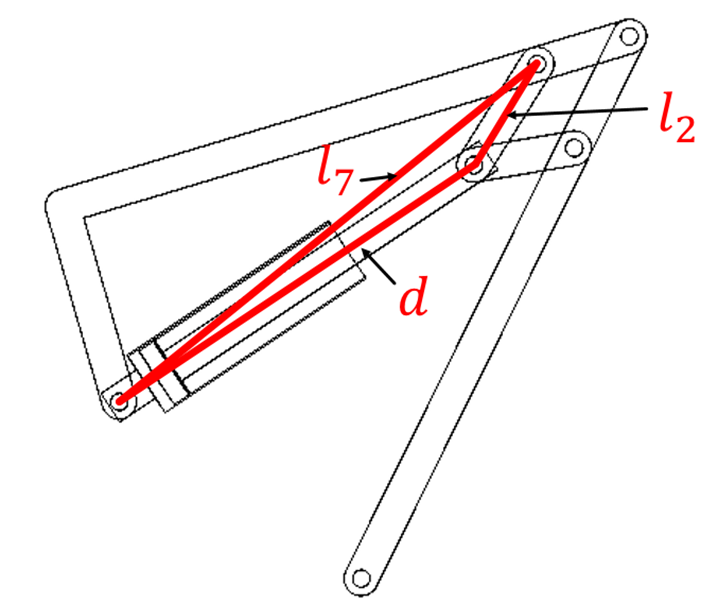}
        \caption{Knee Exoskeleton diagram during Sit position}
        \label{singularity}
    \end{figure}

    \item The optimized dimensions of the exoskeleton can be scaled up or down. To avoid these, variable founds are considered on the basis of leg dimensions of an average human having weight = 80 kg and height = 180 cm : \\  $100 mm > d_1  > 50 mm$ , $100\;mm > d_2  > 50 \; mm $ , $100 \; mm > d_3  > 50 \; mm$, $100 \; mm > d_4  > 50 \; mm$, $ 120 \; mm > d_5  > 50 \; mm $, $300 \; mm > d_6  > 200 \; mm $ . 
\end{enumerate}

\section{Optimization Method and Results}
\begin{figure}[h]
    \centering
    \includegraphics[width =  \textwidth]{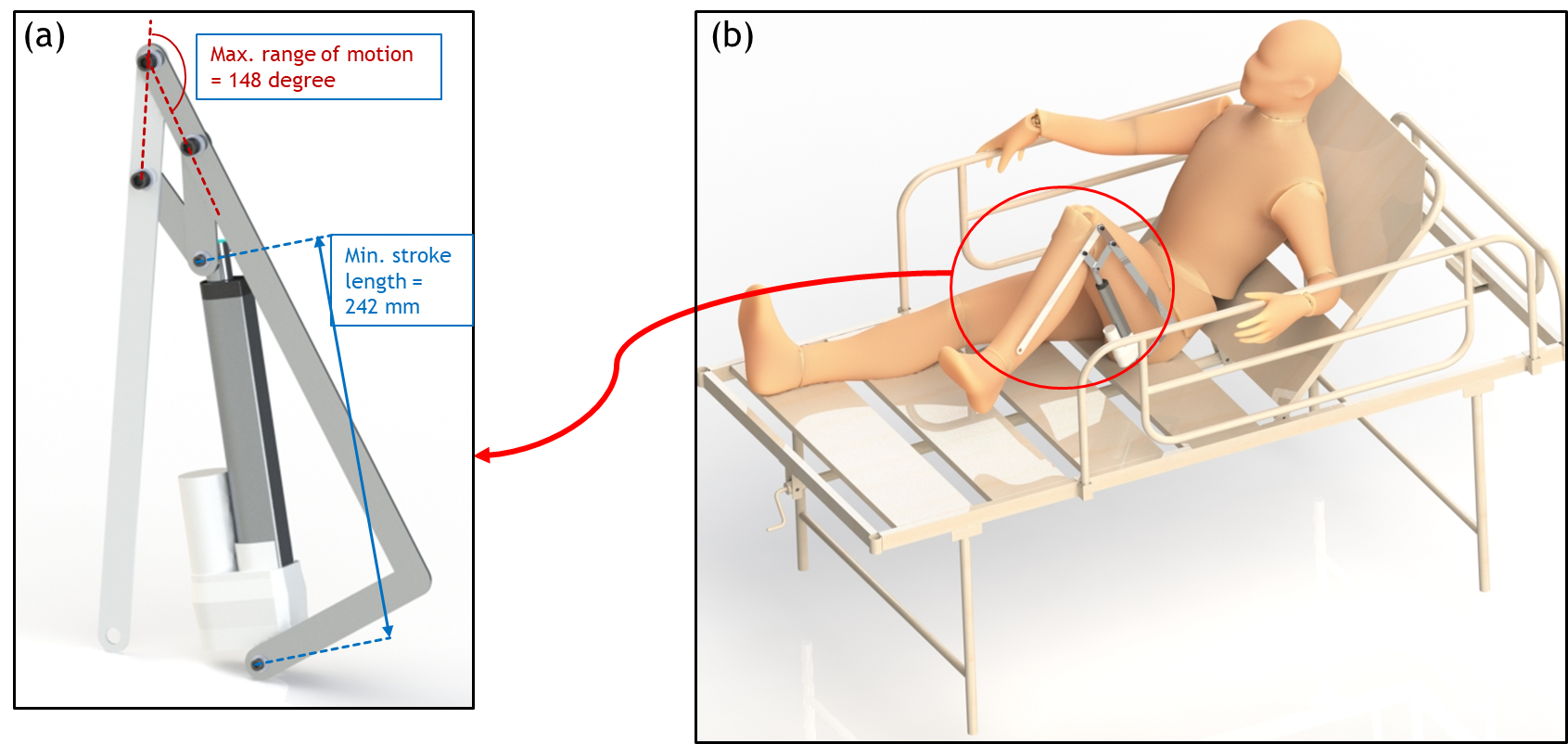}
    \caption{(a) CAD model of optimized exoskeleton (b) Mannequin wearing exoskeleton on hospital bed}
    \label{fig4}
\end{figure}

 The proposed exoskeleton comprises of 6 links, and to obtain the optimal measurement for maximum range of motion for every user, synthesis of exoskeleton has been performed. Interior point method is considered to solve this nonlinear inequality constrained optimization problem. The interior point method which is type of barrier method doesn’t depend on the type of problem. i.e. it can be used for both for convex or non-convex problem. The optimal solution based on this method is given in table \ref{tab:optimum_link_dimensions}. Based on the optimum result, the CAD model is prepared with mannequin to check the feasibility of knee exoskeleton and its maximum range of rotation as shown in Fig \ref{fig4}.

\begin{table}[h]
    \small\sf\centering
    \caption{Optimum link dimensions}
    \label{tab:optimum_link_dimensions}
    \begin{tabular}{ccc}
        \toprule
        \textbf{Design Variable} & \textbf{Initial Value (mm)} & \textbf{Optimum Value (mm)} \\
        \midrule
        $l_1$ & 85  & 59.081 \\
        $l_2$ & 85  & 68.840 \\
        $l_3$ & 85  & 55.964 \\
        $l_4$ & 80  & 71.849 \\
        $l_5$ & 80  & 118.630 \\
        $l_6$ & 235 & 287.310 \\
        \bottomrule
    \end{tabular}
\end{table}

\begin{figure}[H]
    \centering
    \begin{subfigure}{0.45\textwidth}
        \centering
        \includegraphics[width=\textwidth]{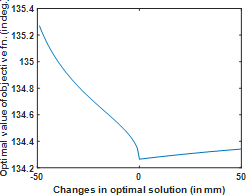}
        \caption{Local minima at optimal solution.}
        \label{optimal graph}
    \end{subfigure}
    \hfill
    \begin{subfigure}{0.45\textwidth}
        \centering
        \includegraphics[width=\textwidth]{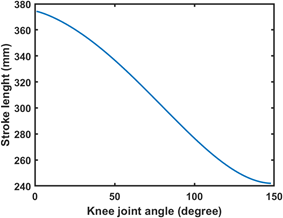}
        \caption{Knee joint angle vs Stroke length.}
        \label{knee joint vs stroke length}
    \end{subfigure}
    \caption{(a) Existence of local minima at optimal solution. (b) Variation of knee joint angle with respect to linear actuator's stroke length.}
    \label{fig:comparison}
\end{figure}

To optimize a given function, it is essential to identify both the optimal point and its corresponding minimum value. This is typically achieved by perturbing the optimal point and analyzing the resulting changes in the solution. As an illustration, a Fig \ref{optimal graph} presents the function's local minima, showing that the objective values increase as one moves away from the optimal point, which is located at the center of the x-axis.
In the Fig \ref{knee joint vs stroke length}, the dependency of stroke length on the knee joint angle has been shown. This graph shows the non-linear relationship between the stroke length and knee joint angle.



\section{Graphical Simulation}
\begin{figure}[H]
    \centering
    \begin{subfigure}{0.45\textwidth}
        \centering
        \includegraphics[width =  \textwidth]{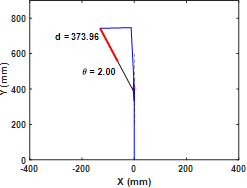}
        \caption{ $\theta$ = $2^{\circ}$ }
        \label{G1}
    \end{subfigure}
    \hfill
    \begin{subfigure}{0.45\textwidth}
        \centering
        \includegraphics[width =  \textwidth]{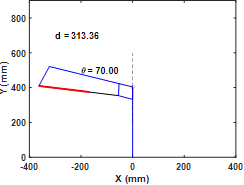}
        \caption{ $\theta$ = $70^{\circ}$ }
        \label{G2}
    \end{subfigure}

    \begin{subfigure}{0.45\textwidth}
        \centering
        \includegraphics[width =  \textwidth]{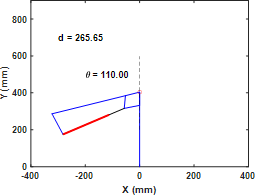}
        \caption{ $\theta$ = $110^{\circ}$ }
        \label{G3}
    \end{subfigure}
    \hfill
    \begin{subfigure}{0.45\textwidth}
        \centering
        \includegraphics[width =  \textwidth]{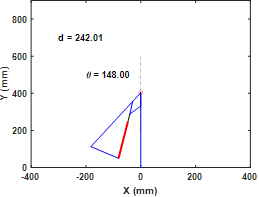}
        \caption{ $\theta$ = $148^{\circ}$ }
        \label{G4}
    \end{subfigure}
    \caption{ Graphical Simulation of optimal exoskeleton at (a) $\theta$ = $2^{\circ}$ (b) $\theta$ = $70^{\circ}$ (c) $\theta$ = $110^{\circ}$ (d) $\theta$ = $148^{\circ}$ }
    \label{graphical sim}
\end{figure}

In constrained optimization problem, the global solution may not necessarily be the desired solution. The global minima might not be feasible in the context of design of the exoskeleton. Therefore, authors have constructed an algorithm to check the feasibility of the solution by providing a geometrical simulation of the knee exoskeleton based on optimal dimension as shown in Fig \ref{graphical sim}. The obtained feasible solution based on the algorithm, is an approximate solution which is a local minima, as shown in the Fig  \ref{optimal graph}.

 In the  Fig \ref{graphical sim}, the ordinate and abscissa represents the dimensions of the links. In Fig \ref{G1}, there is a  rotation of $2^\circ$  i.e. the human is almost standing. As the human start siting, the angle of rotation ($\theta$) increases as shown in Figs \ref{G2} and \ref{G3}. The Fig \ref{G4} shows the maximum angle achieved by the exoskeleton ( $\theta$ = $148^ {\circ}$ ) before reaching the singularity position. The rotation of exoskeleton is actuated by linear actuator as shown in Figs \ref{G1} to \ref{G4}. The thick red line represent the cylinder of the linear actuator and black line represents piston. The combine length of piston and cylinder is represented as $d$ in the Figs. The Fig \ref{G4} validate that the optimal design of exoskeleton with same stroke length ($d_{min}$) gives more range of rotation than inspired design \cite{jain2022linear}. 
The above simulation shows the feasibility of the optimal dimension. The range of motion achieved by this method is 148 degrees which is 24.7$\%$ more than the previous work without optimization on this design. This comparison has been done on the basis of same stroke length. It means that the maximum distance (242 mm) covered by linear actuator is same in both cases.

\section{Gait Analysis}
\begin{figure}[h]
    \centering
    \begin{subfigure}{0.65\textwidth}
        \centering
        \includegraphics[width= 0.4 \linewidth]{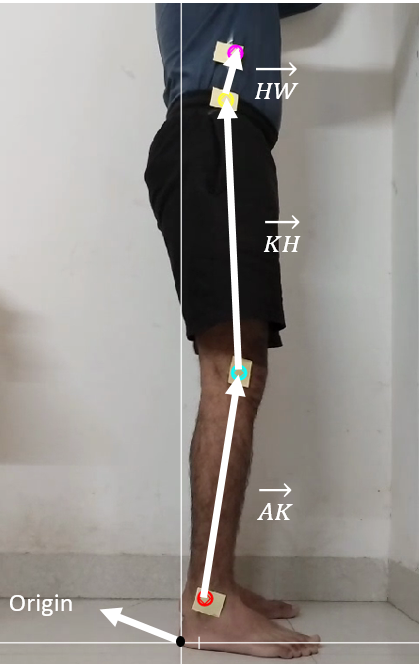}
        \caption{Vector Allocation}
        \label{vec}
    \end{subfigure}
    
    \begin{subfigure}{ \textwidth}
        \centering
        \includegraphics[width=  \linewidth]{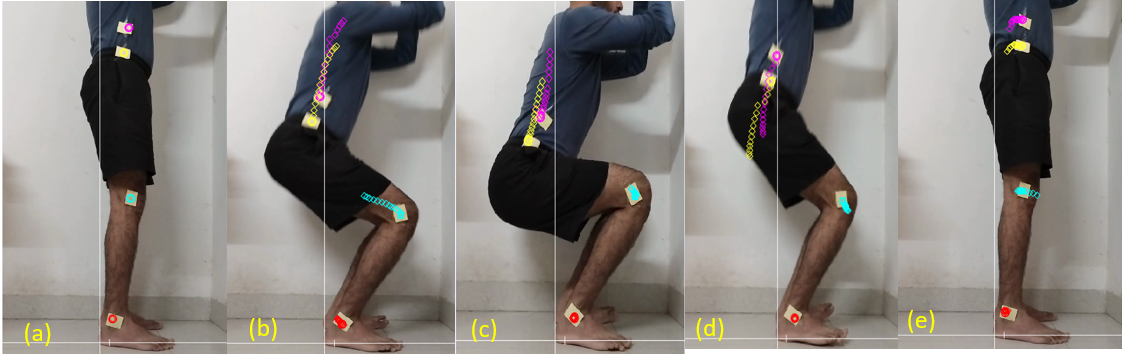}
        \caption{Experiment with human (alone) during STS motion}
        \label{STS}
    \end{subfigure}

    \caption{ Optimum dimension based knee exoskeleton }
    \label{human_sts}
\end{figure}

This section provides the misalignment study of proposed Linear Actuator based Knee Exoskeleton (LAKE) for sit to stand (STS) motion. This can be done by analyzing the human knee's STS motion and LAKE's STS motion.
To perform the gait analysis of human being for STS motion, an experiment is conducted in which subject is performed sit to stand motion. This motion has captured with the help of a high-resolution camera as shown in Fig \ref{STS}. A reflective marker is strategically placed on the knee joint of the subject to ensure precise tracking. The TRACKER software processed the video footage frame by frame, extracting the trajectory data of the marker over time. To analyze the motion of these coordinates, the angles has to be calculated.
 This was done by defining vectors between key anatomical landmarks: $\vec{AK}$ (ankle to knee), $\vec{KH}$ (knee to hip), and $\vec{HW}$ (hip to waist) as shown in Fig \ref{vec}. These vectors were used to determine the angular displacements and provide a comprehensive view of the joint mechanics. The knee joint angle is calculated as
 \begin{align}
     \cos{\theta} = \frac{\vec{AK} \cdot \vec{KH}}{|\vec{AK}| |\vec{KH}|} 
 \end{align}
where, $\theta$ is the knee joint angle.
\begin{figure}[h]
    \centering
    \includegraphics[width=\linewidth]{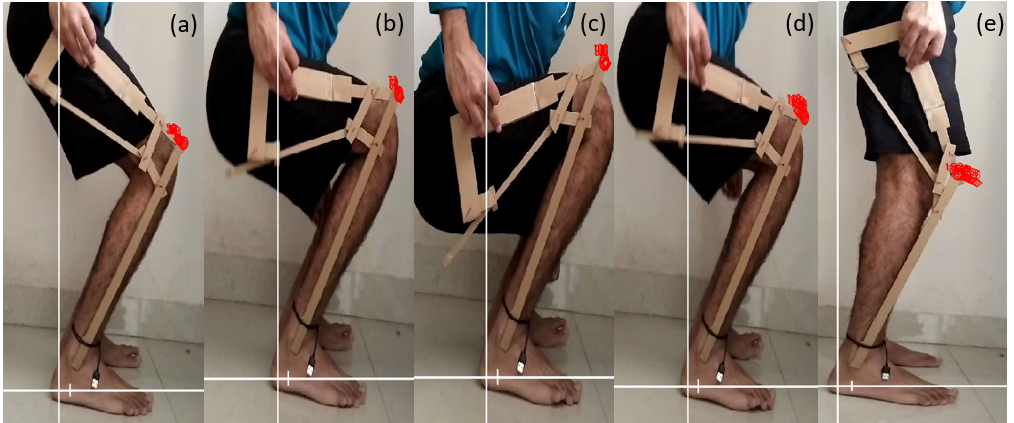}
    \caption{Sit to Stand motion of exoskeleton}
    \label{STS_exo}
\end{figure}

\begin{figure}[h]
    \centering
    \begin{subfigure}{ 0.45 \textwidth}
        \centering
        \includegraphics[width= \linewidth]{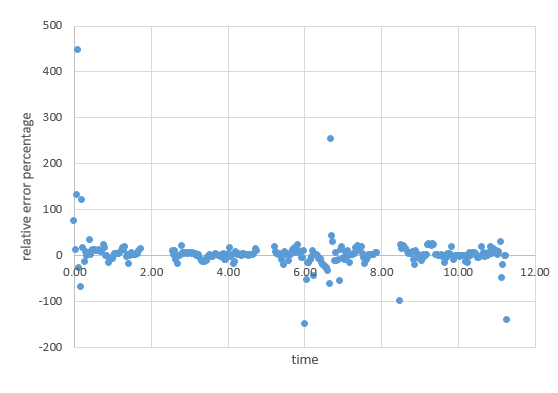}
        \caption{Relative error between both trajectory}
        \label{er}
    \end{subfigure}
    \hfill
    \begin{subfigure}{ 0.45\textwidth}
        \centering
        \includegraphics[width= \linewidth]{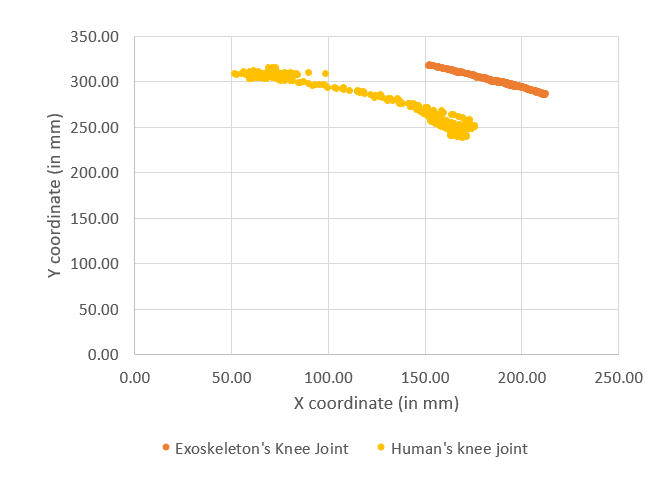}
        \caption{Comparison of the STS motion with and without exoskeleton}
        \label{mot}
    \end{subfigure}

    \caption{ Gait Analysis (a) Relative error between two trajectories (b) Trajectory of human knee adn exoskeleton knee joint.}
    \label{graph}
\end{figure}

Similarly, same subject has performed the STS motion with exoskeleton. This exoskeleton is a dummy model made up from wooden cardboard. Its purpose is to get the trajectory of optimized dimension based knee exoskeleton. In the Fig \ref{STS_exo}, subject is performing sit to stand motion with exoskeleton.
The data obtained from the experiment is used to draw the trajectory of the human's knee joint and exoskeleton's knee joint. The concept behind these generation of trajectories are to get some idea on misalignment between them. The yellow line trajectory shows the motion of knee joint during STS motion as shown in Fig \ref{mot}. The orange line shows the trajectory of the exoskeleton's knee joint. The Figure \ref{er} shows the relative error between both trajectories. Since, both trajectory are similar, the relative error is found as zero at most of the positions. This similarity conclude that the misalignment is minimum between the subject and exoskeleton.
The sudden spike in the error as shown in Fig \ref{er} is due to the human error during the experiment.

\section{Prototype Development}
\begin{figure}[h]
    \centering
    \includegraphics[width=\linewidth]{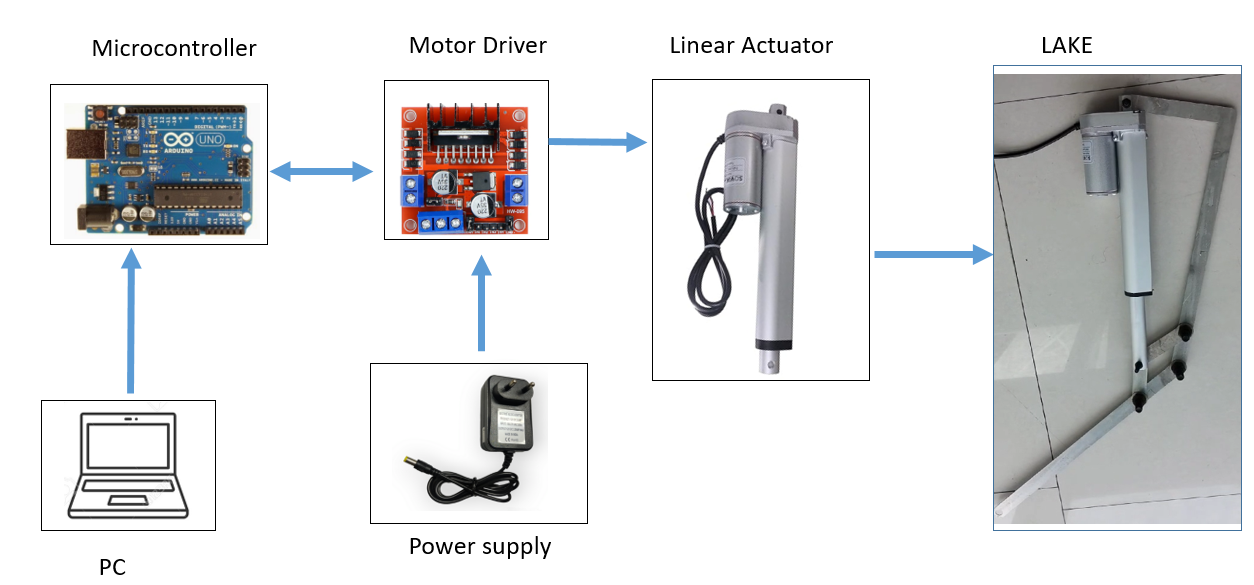}
    \caption{Hardware connectivity}
    \label{hr}
\end{figure}

The prototype of Linear Actuator based Knee joint Exoskeleton (LAKE) has been developed. In this work, the micro-controller used is Arduino UNO, as shown in Fig \ref{hr}. 
The power supply has been provided by laptop. To control the linear actuator, the motor driver L298N has been used. This is important because microcontrollers often cannot supply the necessary current to drive motors directly. Motor drivers amplify the control signals from the microcontroller to provide the appropriate power to the motors.The author used L298N motor driver because of its high current capability, as the L298N can drive motors with up to 2A of current per channel, accommodating a broad range of motor types and sizes. The module operates over a wide voltage range, handling motor supply voltages from 5V to 35V, which adds to its adaptability in various projects.
Speed control is another critical feature of the L298N, achieved through pulse-width modulation (PWM). By varying the duty cycle of the PWM signal applied to the enable pins (ENA and ENB), users can finely adjust motor speeds. 

\begin{figure}[h]
    \centering
    \includegraphics[width=0.6\linewidth]{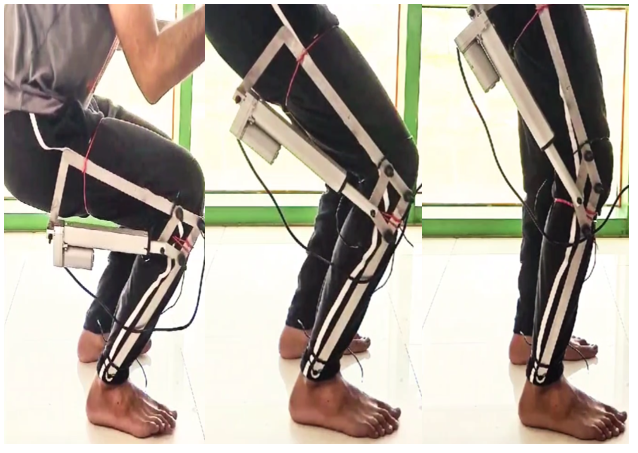}
    \caption{Prototype of LAKE with STS motion}
    \label{proto}
\end{figure}

\begin{figure}
    \centering
    \includegraphics[width=\linewidth]{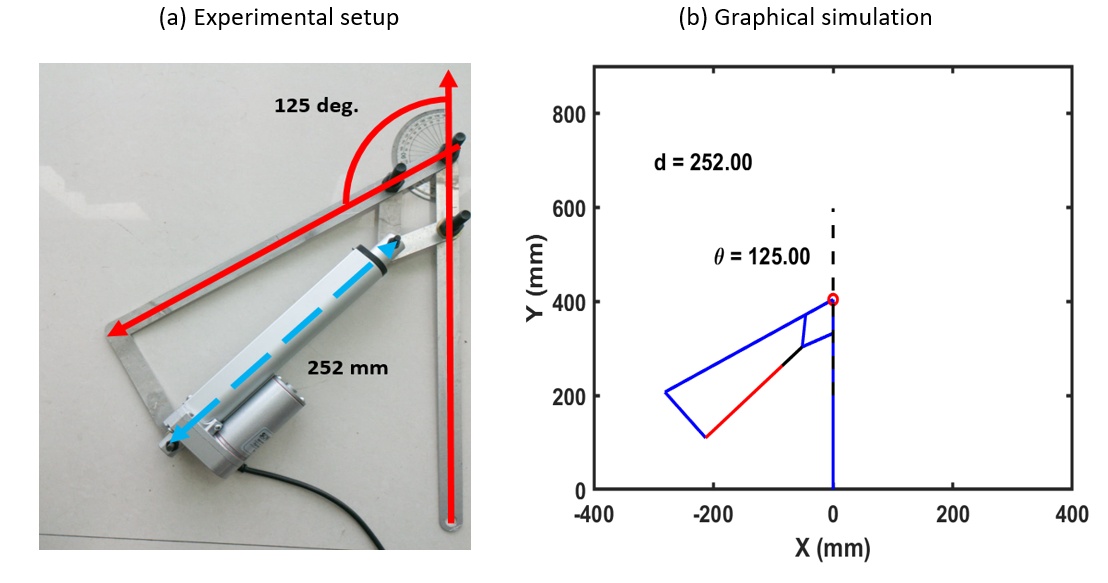}
    \caption{Validation of Graphical simulation with experimental results.}
    \label{valid}
\end{figure}

After assembling all the components shown in Fig \ref{hr}, the C language based code is written in Arduino IDE. This code is uploaded to the Arduino which drives the motor driver and further the exoskeleton as shown in Fig \ref{proto}. The exoskeleton worn by the subject in Fig \ref{proto} operates at a stroke velocity of $5 \;mm/s $. The subject receives no external support other than the exoskeleton itself. The upward motion is facilitated solely by the forces generated through actuator actuation.
Furthermore, this prototype has been used to validate the graphical simulation as shown  in Fig \ref{graphical sim} . The angle achieved by the LAKE in both the graphical simulation and the physical prototype is found to be equal at the given stroke length, as shown in the Fig \ref{valid}.

\section{ Conclusion}
In this paper, the optimization algorithm has been constructed which optimize the design of closed loop 4-bar mechanism knee exoskeleton for maximum range of rotation. To achieve that, the algorithm used the nonlinear equation derived from kinematic modeling of exoskeleton and optimized it using interior point method and then check the feasibility of result by simultaneously constructing the geometrical simulation of the optimized knee exoskeleton for maximum range of rotation. This optimized exoskeleton has been modeled using CAD software to further validate its maximum range of motion and feasibility by harnessing on the human mannequin based on average human measurement.
To investigate joint misalignment, an experiment was conducted in two stages: first, with a human knee joint performing the sit-to-stand (STS) motion, and then with a dummy fitted with the optimized knee exoskeleton. The marker on the knee joint and software provide us the trajectory of the human knee joint and knee exoskeleton joint. The relative error between the two trajectories ensured that knee exoskeleton joint and human joint are almost in sync during STS motion. 
Furthermore, to validate the optimized results, a prototype of Linear actuator based Knee exoskeleton (LAKE) was developed. The prototype utilizes an Arduino microcontroller and operates under an open-loop control scheme to perform the STS motion.

\clearpage
\newpage
\bibliographystyle{unsrt}  
\bibliography{references}

\end{document}